\documentclass{article}

\usepackage[preprint]{neurips_2024}

\usepackage[utf8]{inputenc} 
\usepackage[T1]{fontenc}    
\usepackage{hyperref}       
\usepackage{url}            
\usepackage{booktabs}       
\usepackage{amsfonts}       
\usepackage{nicefrac}       
\usepackage{microtype}      
\usepackage{xcolor}         
\usepackage{graphicx}
\usepackage{amsmath}
\usepackage[linesnumbered,ruled,vlined]{algorithm2e}
\usepackage{subcaption}
\usepackage{caption}

\title{Multimodal Representation Learning using Adaptive Graph Construction}

\author{Weichen Huang \\
Department of Computer Science\\
St Andrew's College\\
Dublin, A94XN72, Ireland \\
\texttt{\{w.huang\}@students.st-andrews.ie} \\
}

\begin{document}

\maketitle

\begin{abstract}
  Multimodal contrastive learning train neural networks by levergaing data from heterogeneous sources such as images and text. Yet, many current multimodal learning architectures cannot generalize to an arbitrary number of modalities and need to be hand-constructed. We propose AutoBIND, a novel contrastive learning framework that can learn representations from an arbitrary number of modalites through graph optimization. We evaluate AutoBIND on Alzhiemer's disease detection because it has real-world medical applicability and it contains a broad range of data modalities. We show that AutoBIND outperforms previous methods on this task, highlighting the generalizablility of the approach.
\end{abstract}

\section{Introduction}

The human brain possesses a remarkable capacity to learn and integrate information from diverse sensory modalities in a dynamic manner. This is similar to the challenge of multimodal contrastive learning. Just as the brain synthesizes inputs from various senses like images, sound, text and numbers to form a cohesive understanding of the world, multimodal contrastive learning aims to train neural networks by leveraging data from different sources. Among existing multimodal learning approaches, several methods have made significant strides. ~\cite{hager2023best} proposes a method for handling tabular and image data through the combination of self-supervised learning and contrastive learning. However, it exhibits limitations when confronted with more than two modalities, and furthermore, its approach exclusively supports single modality outputs, thus constraining its versatility.





ImageBIND~\cite{girdhar2023imagebind}, specializes in binding multiple modalities but within a fixed number. This technique utilizes image data to harmonize diverse modalities. However, its weakness emerges when confronted with missing image data, rendering it less robust in such scenarios. ~\cite{huang2023multimodal} presents the latest framework for multimodal contrastive learning of medical images and tabular data, applying the techniques to Alzheimer's disease prediction. With carefully crafted contrastive neural topologies, it claims over 83.8\% prediction accuracy, 10\% increase from the state of the art solutions.

Unfortunately, all these existing multimodal contrastive learning approaches, although promising, fall short in achieving generalization across an arbitrary number of modalities. They demand meticulous manual construction and often prove vulnerable when dealing with absent modalities. These methods are inherently shaped by the specific knowledge and assumptions associated with the modalities they address, making them less adaptable to scenarios requiring universality. To the best of our knowledge, no prior work in multimodal contrastive learning has proposed solutions to tackle the challenge of learning with an arbitrary number of modalities in a universally applicable manner.

In this work, we present \textbf{AutoBIND}, a versatile and universally applicable framework for contrastive learning with an arbitrary number of modalities. To effectively learn and refine the representations of each modality, AutoBIND employs contrastive loss that highlights the nuanced relationships between the data sources. A distinguishing feature of AutoBIND is its dynamic graph structure adaptation during training. This adaptive graph mechanism is a pivotal solution for handling the absence of certain modalities, an issue that often plagues traditional multimodal learning techniques. Just as the human brain adapts to the absence of certain sensory inputs by amplifying the significance of others, AutoBIND dynamically updates the graph's topology to effectively accommodate and learn from the available modalities. Moreover, this framework improves the model's ability to handle heterogeneous data arrangements: cases in which some modalities are available while others are not.

\section{Problem Setup}

%




In our setup, we consider aribitary $n$ modalities, where each modality $i$ has a set of instances $X_i$ and an encoder function $f_i$ that learns representations $Z_i$ for each modality. Our goal is to learn a shared embedding space where representations of similar instances across different modalities are brought closer together, while representations of dissimilar instances are pushed apart. We represent the problem of multimodal constrastive learning as an undirected graph $G = (V, E)$, where $V$ is the set of nodes, each corresponding to a modality $i$, and $E$ is the set of edges, where an edge $(i, j)$ represents the correlation between modality $i$ and modality $j$. For each edge $(i, j) \in E$, we can define a similarity function $\text{Sim}(Z_i, Z_j)$ using cosine similarity to express the correlation between modalities $i$ and $j$ in the shared embedding space. Thus, we define the distance between two nodes as $d_{ij} \propto \frac{1}{\text{Sim}(Z_i, Z_j)}$.

The optimization objective involves minimizing the contrastive loss, which can be expressed using the similarity and dissimilarity functions as follows:

\begin{equation}
    \mathcal{L}(Z_i, Z_j) = -\log\left( \frac{\exp(\text{Sim}(Z_i, Z_j))}{\exp(\text{Sim}(Z_i, Z_j)) + \sum_{k \neq i} \exp(\text{Dissim}(Z_i, Z_k))} \right)
\end{equation}

Now, let's consider two sets of modalities $i$ (correlated modalities) and $j$ (uncorrelated modalities). The argument states that arranging correlated modalities ($i$) together in the graph leads to a lower overall loss than mixing them with uncorrelated modalities ($j$): $\sum_{m \in i} \mathcal{L}(Z_m, Z_m') + \sum_{n \in j} \mathcal{L}(Z_n, Z_n') < \sum_{p \in i \cup j} \mathcal{L}(Z_p, Z_p')$, shere $Z_m$ and $Z_m'$ are representations of correlated modalities $m$, and $Z_n$ and $Z_n'$ are representations of uncorrelated modalities $n$. $Z_p$ and $Z_p'$ are representations of modalities in the combined set $i \cup j$. At the end of the graph optimization process, where nodes representing correlated modalities are strategically positioned closer together and uncorrelated modalities are separated, this arrangement directly influences the behavior of the original contrastive loss $\mathcal{L}$ to be minimized.

\section{Proposed Methods}

\subsection{Graph Construction}

We consider two different approaches to constructing the graph: fully connected graph (FCG) and minimum spanning tree (MST). The fully connected graph is the simplest approach, where each modality is connected to every other modality. The minimum spanning tree is a tree-based approach that selects the most correlated modalities and connects them together. 

\textbf{Fully Connected Graph.} A fully connected graph (FCG) is selected as the initial representation, denoted by $G_{\text{full}} = (V, E_{\text{full}})$, where each modality $i$ corresponds to a node in $V$, and the edge set $E_{\text{full}}$ includes all possible edges $(i, j)$ between modalities. This choice is made to comprehensively capture modalities' intricate correlations within the contrastive learning framework.

\textbf{Minimum Spanning Tree.} The adoption of a minimum spanning tree (MST), achieved through Kruskal's algorithm is motivated by the opposing theory of removing noise and redundancy, leading to a more interpretable framework. The MST, denoted as $G_{\text{MST}} = (V, E_{\text{MST}})$, is a subgraph of the FC graph that retains the fundamental correlations while eliminating excess edges. Based on the derived edge weights, we prune the nodes with lowest sum of the edge weights.


\subsection{Graph Update}

In each iteration ($batch\_num$), the algorithm follows these steps:

\begin{algorithm}[H]
   \SetAlgoLined
   \KwData{List of modalities}
   \KwResult{Minimum Spanning Tree $G_{\text{MST}}$}
   Initialize an empty list of edges $E_{\text{empty}}$\;
   \For{each pair of modalities $i$ and $j$}{
       Calculate the correlation factor between $i$ and $j$ and store it in $E_{\text{empty}}$\;
   }
   Sort $E_{\text{empty}}$ in non-decreasing order of correlation weights\;
   Form an empty graph $G_{\text{MST}}$ representing the minimum spanning tree\;
   \ForEach{sorted edge $(u, v)$}{
       \If{adding edge $(u, v)$ to $G_{\text{MST}}$ does not create a cycle}{
           Incorporate edge $(u, v)$ into $G_{\text{MST}}$\;
           Merge disjoint sets of nodes $u$ and $v$\;
           Train the contrastive model on modality pair $(u, v)$ using get\_embedding($u$, batch\_num) and get\_embedding($v$, batch\_num)\;
       }
   }

   \caption{Multimodal Minimum Spanning Tree (MMST) Algorithm}
\end{algorithm}

   

\begin{figure*}[h]
    \centering
    \includegraphics[width=0.7\textwidth]{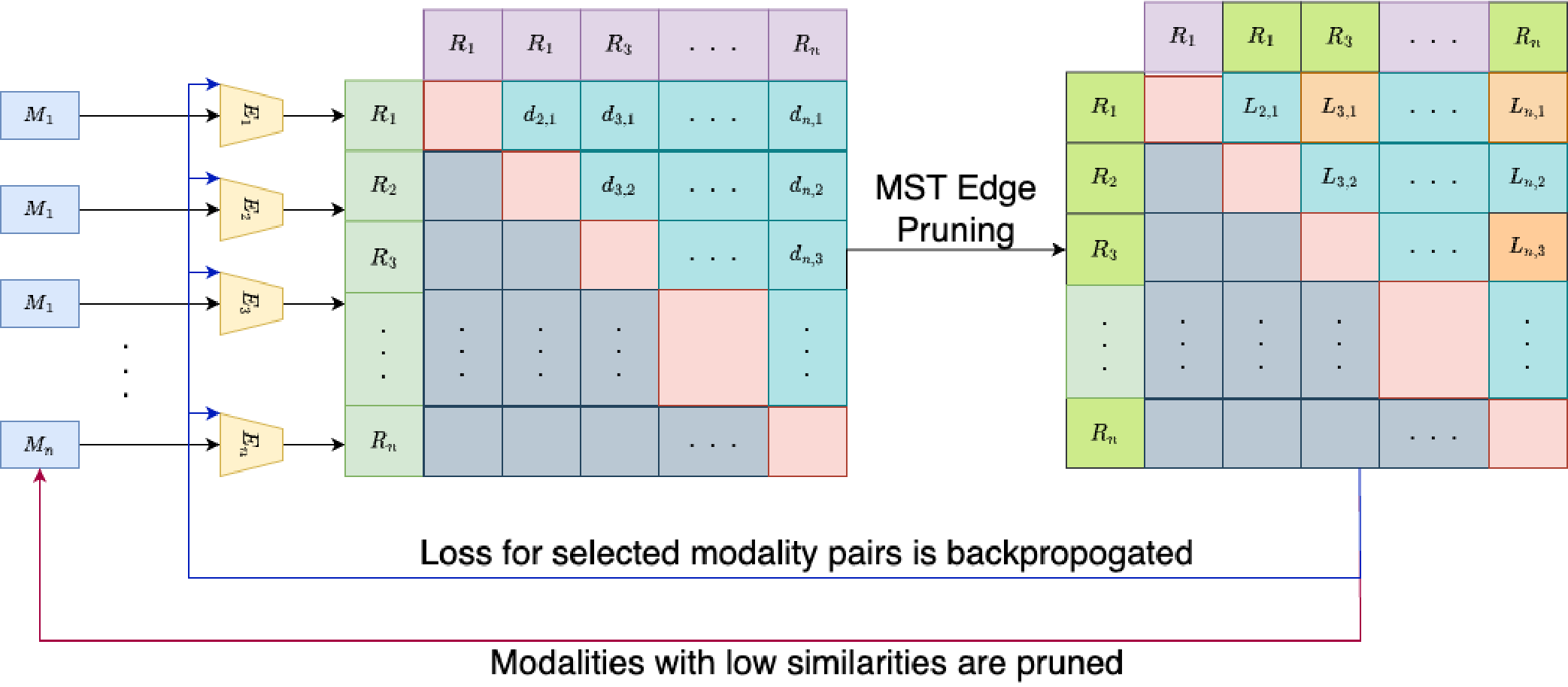}
    \caption{Overview of the AutoBIND Process: Illustration depicting the various stages and steps involved in the AutoBIND framework. The process encompasses multimodal embedding and graph construction, resulting in enhanced performance across different datasets.}
    \label{fig:autobind}
\end{figure*}

\section{Experiments}

\subsection{Dataset}

Our evaluation methodology unfolds through a series of strategic steps. Firstly, we encode each modality while encompassing every label class. Subsequently, the embeddings originating from each modality are thoughtfully concatenated. Leveraging cosine similarity, we pinpoint the class exhibiting the highest similarity score. We use typical classification metrics to evaluate the model: accuracy, precision, recall.

We utilize the Alzheimer's Disease Neuroimaging Initiative (ADNI) dataset ~\cite{jack2008alzheimer}, a comprehensive collection of multimodal data encompassing subjects with a spectrum of cognitive states, including normal cognition, mild cognitive impairment (MCI), and Alzheimer's disease (AD). The ADNI dataset comprises diverse modalities, including tabular data and medical images. Furthermore, the ADNI dataset contains missing tabular values in the tabular data, making it an ideal candidate for evaluating our model's robustness on heterogeneous data arrangements.

We compare our approaches with three baselines: a 2D ResNet baseline~\cite{sun2021improved} and a 3D CNN baseline~\cite{song2021effective} and MedBIND~\cite{huang2023multimodal}.

\subsection{Results}

\begin{figure}[ht]
    \centering

    \begin{subfigure}{0.3\textwidth}
        \includegraphics[width=\linewidth]{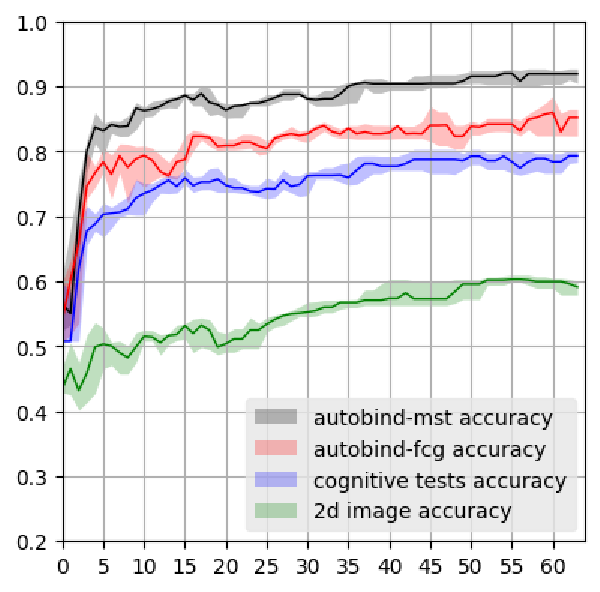}
        \caption{Accuracy} 
        \label{fig:perf_accuracy}
    \end{subfigure}\hspace*{\fill}
    \begin{subfigure}{0.3\textwidth}
        \includegraphics[width=\linewidth]{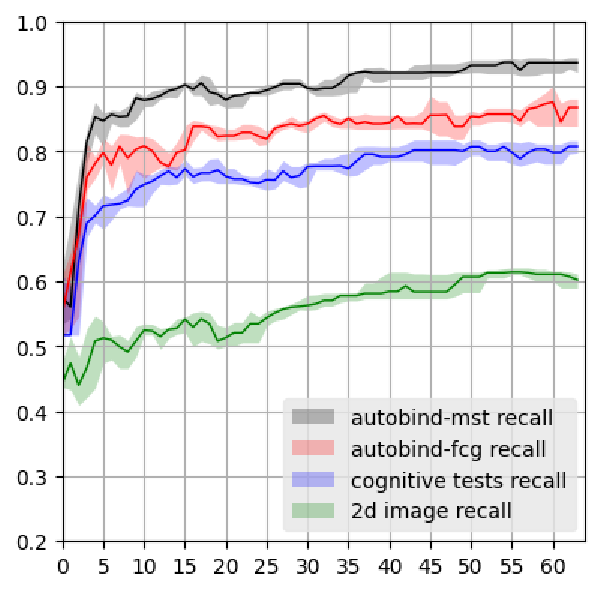}
        \caption{Recall} 
        \label{fig:perf_recall}
    \end{subfigure}\hspace*{\fill}
    \begin{subfigure}{0.3\textwidth}
        \includegraphics[width=\linewidth]{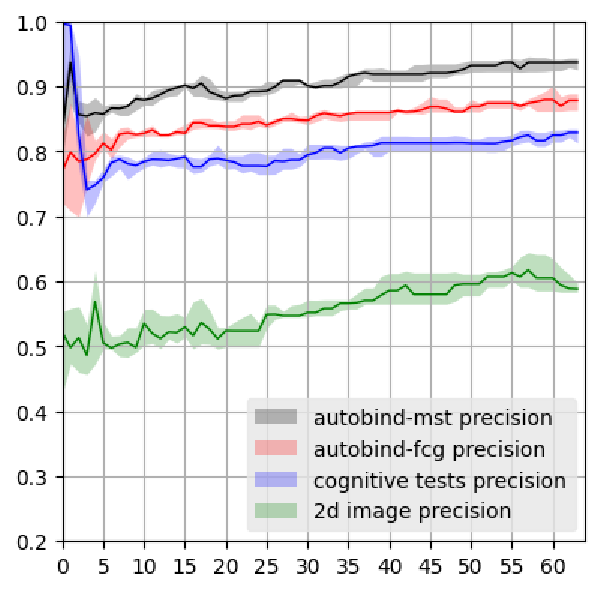}
        \caption{Precision} 
        \label{fig:perf_precision}
    \end{subfigure}
    
    \caption{Performance of AutoBIND MST vs. AutoBIND FCG vs Unimodals.}
    \label{fig:perf_compare}
\end{figure}

\begin{table}[t]    
    \vskip 0.15in
    \begin{center}
    \begin{small}
    \begin{sc}
        \resizebox{\textwidth}{!}{\begin{tabular}{llccccccc}
        Model & Modality & Accuracy & Precision & Recall & M-M & APT & RES & T-A \\
        \hline
        2D ResNet & 2D MRI & 0.799 $\pm$ 0.000 & 0.770 $\pm$ 0.124 & 0.799 $\pm$ 0.000 & - & - & - & - \\
        3D CNN    & 3D MRI-PET & 0.745 $\pm$ 0.064 & 0.594 $\pm$ 0.082 & 0.799 $\pm$ 0.000 & - & - & - & - \\
        MedBIND   & 2D MRI & 0.838 $\pm$ 0.023 & NA $\pm$ NA & 0.799 $\pm$ 0.000 & \checkmark & - & - & \checkmark \\
        \hline
        AutoBIND MST & Multi & \textbf{0.916 $\pm$ 0.014} & \textbf{0.936 $\pm$ 0.012} & \textbf{0.933 $\pm$ 0.014} & \textbf{\checkmark} & \textbf{\checkmark} & \textbf{\checkmark} & \textbf{\checkmark} \\
        AutoBIND FCG & Multi & 0.848 $\pm$ 0.028 & 0.877 $\pm$ 0.020 & 0.864 $\pm$ 0.028 & \checkmark & \checkmark & - & \checkmark \\
    \end{tabular}}
    \caption{Performance for AutoBIND on ADNI Datasets\\(M-M: multimodal, APT: adaptability, RES: resilience to missing data, T-A: task agnostic)}
    \label{tab:models}
    \end{sc}
    \end{small}
    \end{center}
    \vskip -0.1in
\end{table}

From Table~\ref{tab:models}, we can see that AutoBIND MST outperforms the baseline models and MedBIND in terms of accuracy, precision and recall. This shows that the MST graph construction method and node pruning is more effective than the FC graph construction method. Furthermore, AutoBIND FCG also outperforms the baseline models and MedBIND in terms of accuracy, precision and recall. This illustrates the importance of multimodal contrastive learning in improving performance.

Compared to existing works, both AutoBIND graph construction methods are task agnostic, meaning that they can be applied to any task in any domain. Furthermore, they are also adaptable to different datasets and encoders, whereas existing models are task-specific and rely on static frameworks for multimodal learning. Finally, our models are also resilient to missing data, as the tabular data in the ADNI dataset contains missing values.

\section{Discussion}

In conclusion, our endeavor has yielded a novel approach to multimodal contrastive learning, unveiling a methodology that surmounts numerous challenges to deliver noteworthy outcomes across a spectrum of datasets. Demonstrating the prowess of our approach, we have showcased its superior performance on a real-world medical dataset. Specifically, it surpasses existing methods such as ImageBIND on the ADNI dataset. Notably, the orchestrated graph structure, a cornerstone of our methodology, converges to an optimal framework for multimodal learning. This adaptively evolving structure dynamically adapts to the available modalities, effectively enhancing learning and accommodating variable data configurations. 

Future works include an investigation of alternative graph structures that could further refine our model's performance. Our curiosity extends to uncharted modalities such as audio and video, aiming to harness the power of our approach in realms beyond text, images, and tables. Indeed, future works must validate this framework on a wider variety of data modalities to ensure its generalizablility. Moreover, there is still more work to be done on the framework itself. Since the two proposed graph constructions are only heuristically determined due to the exponential complexity of iterating through the full search space, there remains work in proving its optimality or finding a better construction.


\bibliography{neurips_hs_2024}
\bibliographystyle{iclr2024_conference}

\end{document}